\documentclass{article}

\PassOptionsToPackage{numbers, compress}{natbib}

\usepackage[final]{nips_2018}

\usepackage[utf8]{inputenc} 
\usepackage[T1]{fontenc}    
\usepackage{hyperref}       
\usepackage{url}            
\usepackage{booktabs}       
\usepackage{amsfonts}       
\usepackage{nicefrac}       
\usepackage{microtype}      
\usepackage{xcolor}
\usepackage{parsetree}
\usepackage{subcaption}

\title{Improving Semantic Parsing for Task Oriented Dialog}

\author{
  Arash Einolghozati\thanks{equal contribution} \\
  Facebook Conversational AI\\
  \texttt{arashe@fb.com} \\
   \And
   Panupong Pasupat\footnotemark[1]\\
    Stanford University\\
   \texttt{ppasupat@cs.stanford.edu} \\
   \And
   Sonal Gupta \\
  Facebook Conversational AI\\
   \texttt{sonalgupta@fb.com} \\
   \And
   Rushin Shah \\
     Facebook Conversational AI\\
   \texttt{rushinshah@fb.com} \\
   \And
   Mrinal Mohit \\
  Facebook Conversational AI\\
   \texttt{mrinalmohit@fb.com} \\
      \And
   Mike Lewis \\
  Facebook AI Research\\
   \texttt{mikelewis@fb.com} \\
      \And
  Luke Zettlemoyer  \\
  Facebook AI Research\\
   \texttt{lsz@fb.com} \\
}
\begin{document}

\maketitle

\begin{abstract}
Semantic parsing using hierarchical representations has recently been proposed for task oriented dialog with promising results. 
In this paper, we present three different improvements to the model: contextualized embeddings, ensembling, and pairwise re-ranking based on a language model. We taxonomize the errors possible for the hierarchical representation, such as wrong top intent, missing spans or split spans, and show that the three approaches correct different kinds of errors. The best model combines the three techniques and gives~6.4\% better exact match accuracy than the state-of-the-art, with an error reduction of 33\%, resulting in a new state-of-the-art result on the Task Oriented Parsing (TOP) dataset. 
\end{abstract}

\section{Introduction}
Although intelligent personal assistants such as Alexa and Siri are now ubiquitous, modeling the semantics of compositional natural language queries remains challenging.
The most common practice for natural language understanding in task oriented dialogs is to apply a slot-filling system~\cite{mesnil2013,Liu2016AttentionBasedRN}.
Given a simple query such as \emph{``How far is San Francisco''}, a slot-filling system would classify the \emph{intent} of the whole query (\texttt{GET\_DISTANCE}) and tag the relevant \emph{slots} (\emph{San Francisco} being a \texttt{DESTINATION}) with a sequence tagging model.
However, understanding more complex queries such as \emph{``How far is the coffee shop''}, which requires locating the coffee shop (\texttt{GET\_RESTAURANT\_LOCATION} intent) before finding the distance (\texttt{GET\_DISTANCE} intent), is not straightforward in traditional systems, since the compositionality is not captured when the task is posed as text classification or sequence tagging.

Recently, a Task Oriented Parsing (TOP) representation for intent-slot based dialog systems was introduced ~\cite{gupta2018rnng}. The representation, as illustrated in Figure~\ref{fig:examples}, is expressive enough to capture the task-specific semantics of complex nested queries, but is easier to annotate and parse than full semantic representations such as logical forms \cite{zelle1996learning} or abstract meaning representation \cite{banarescu2013amr}.

A key advantage of the TOP representation is that its structure is similar to standard constituency parses, allowing us to adapt improvements in modeling techniques for syntactic parsers to this problem. In this paper, we propose three approaches to improve the parsing model:

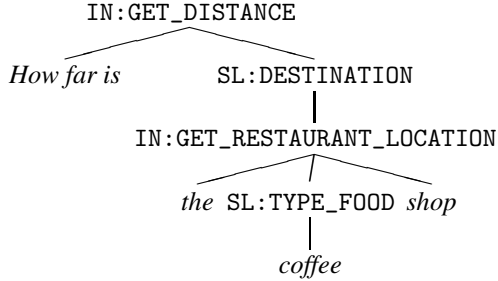
\begin{itemize}
    \item Ensembling with three strategies: majority voting, greedy action, and parser switch.
    \item Incorporating deep contextualized word embeddings, such as ELMo~\cite{Peters:2018}.
    \item Re-ranking the parses based on a language model (LM).
\end{itemize}

To compare the effectiveness of the different approaches,
we also propose an error classification scheme for the TOP representation. 
Our analysis shows that the improvements from the three techniques are orthogonal. 
This allows us to construct the best model using an ensemble of seven models, of which three are trained with LM-based re-ranking.


\section{Base Model and Data}

\begin{figure*}[t]
        
    \centering
        \begin{parsetree}
            \pthorgap{5pt}                         
            \ptvergap{12pt}                        
            \ptnodefont{\normalsize\rm}{9pt}{3pt}  
            \ptleaffont{\normalsize\it}{9pt}{3pt}  
            (.\texttt{IN:GET\_DISTANCE}. `How far is' (.\texttt{SL:DESTINATION}. (.\texttt{IN:GET\_RESTAURANT\_LOCATION}. `the' (.\texttt{SL:TYPE\_FOOD}. `coffee' ) `shop' ) ) )
        \end{parsetree}
    \\[1em]
    \begin{tabular}{cl@{\hspace{-15ex}}r}
    \toprule
    \cmidrule(r){1-3}
    Action & Stack & Buffer \\
    \midrule
    & & \emph{How far is the coffee shop} \\
    \texttt{IN:GD} & \texttt{IN:GD} &  \emph{How far is the coffee shop} \\
    \texttt{SHIFT} & \texttt{IN:GD} \emph{How} &  \emph{far is the coffee shop} \\
    \texttt{SHIFT} & \texttt{IN:GD} \emph{How far} &  \emph{is the coffee shop} \\
    \texttt{SHIFT} & \texttt{IN:GD} \emph{How far is} &  \emph{the coffee shop} \\
    \texttt{SL:DEST} & \texttt{IN:GD} \emph{How far is} \texttt{SL:DEST} &  \emph{the coffee shop} \\
    \texttt{IN:GRL} & \texttt{IN:GD} \emph{How far is} \texttt{SL:DEST} \texttt{IN:GRL} &  \emph{the coffee shop} \\
    \texttt{SHIFT} & \texttt{IN:GD} \emph{How far is} \texttt{SL:DEST} \texttt{IN:GRL} \emph{the} &  \emph{coffee shop} \\
    \texttt{SL:TF} & \texttt{IN:GD} \emph{How far is} \texttt{SL:DEST} \texttt{IN:GRL} \emph{the} \texttt{SL:TF} &  \emph{coffee shop} \\
    \texttt{SHIFT} & \texttt{IN:GD} \emph{How far is} \texttt{SL:DEST} \texttt{IN:GRL} \emph{the} \texttt{SL:TF} \emph{coffee} & \emph{shop} \\
    \texttt{REDUCE} & \texttt{IN:GD} \emph{How far is} \texttt{SL:DEST} \texttt{IN:GRL} \emph{the} \texttt{[SL:TF} \emph{coffee} \texttt{]} & \emph{shop} \\
    \texttt{SHIFT} & \texttt{IN:GD} \emph{How far is} \texttt{SL:DEST} \texttt{IN:GRL} \emph{the} \texttt{[SL:TF} \emph{coffee} \texttt{]} \emph{shop} \\
    \texttt{REDUCE} & \texttt{IN:GD} \emph{How far is} \texttt{SL:DEST} \texttt{[IN:GRL} \emph{the} \texttt{[SL:TF} \emph{coffee} \texttt{]} \emph{shop} \texttt{]} \\
    \texttt{REDUCE} & \texttt{IN:GD} \emph{How far is} \texttt{[SL:DEST} \texttt{[IN:GRL} \emph{the} \texttt{[SL:TF} \emph{coffee} \texttt{]} \emph{shop} \texttt{]} \texttt{]} \\
    \texttt{REDUCE} & \texttt{[IN:GD} \emph{How far is} \texttt{[SL:DEST} \texttt{[IN:GRL} \emph{the} \texttt{[SL:TF} \emph{coffee} \texttt{]} \emph{shop} \texttt{]} \texttt{]} \texttt{]} \\
    \bottomrule
    \end{tabular}
    
    \caption{An example TOP annotation. Intents are prefixed with \texttt{IN:} and slots with \texttt{SL:}.  In a traditional intent-slot system, the \texttt{SL:DESTINATION} could not have an intent nested inside it. The table below shows the parser actions used to generate the parse (with abbreviated labels). At each step, the candidate actions are scored based on the stack LSTM embeddings of three sequences: the previous actions, the current stack, and the current buffer.}
    \label{fig:examples}
\end{figure*}

Following \cite{gupta2018rnng},
we use a shift-reduce parser based on Recurrent Neural Network Grammars (RNNG)~\cite{dyer-rnng:16} as our base model.
The model constructs a parse tree by predicting a sequence of actions. The set of actions include \texttt{SHIFT}, \texttt{REDUCE}, and the generation of intent and slot labels.
The \texttt{SHIFT} action consumes an input token, adding it as a child of most recent `open' sub-tree node, while the \texttt{REDUCE} action closes a sub-tree. The final set of actions is to generate a non-terminal (an intent or a slot) as a new empty sub-tree node.
Note that at each step, only a subset of actions will be valid. For examples, if all the input tokens have been added to the tree, the only valid action is \texttt{REDUCE}. 

In an RNNG model, the action are scored based on the embeddings of three sequences: the list of actions performed so far, the partial construction of the tree (``stack''), and the list of remaining tokens (``buffer''). Each sequence is embedded using a stack LSTM, which dynamically recomputes the embedding as items are added to or removed from the sequence.

For training, we use the same hyperparameters as in~\cite{gupta2018rnng}: 2-layer LSTMs of size 164, dropout rate of 0.34, and pre-trained word embeddings of size 200. We train with 16 workers using Hogwild \cite{recht2011hogwild} for 1 epoch. For optimization, we use Adam \cite{kingma2015adam} with learning rate of 0.0004 and weight decay of 0.00004. The base model uses greedy decoding for inference at test time.

The open-sourced data released by~\cite{gupta2018rnng} contains utterances of navigation and event domains. We remove the utterances where the top intent is \texttt{IN:UNSUPPORTED} as it is a noisy catch-all class for out-of-domain utterances. The final dataset contains 28,276 training, 4,014 evaluation, and 8,191 test utterances. Our evaluation metric is the exact match accuracy: the number of utterances whose full parse trees are correctly predicted.

\section{Improvements}

\subsection{Ensembling}

Ensembling is a powerful method to improve the performance of dependency and constituency parsers \cite{Henderson99Ensemble,Surdeanu2010ensemble}.
In order to take advantage of parser diversity (e.g., sentence shuffling, dropout seeds, parameter initialization), we use different strategies to combine the output of individual parsers. 

A parse tree has a one-to-one correspondence with the sequence of actions performed by the parser. 
In the approaches below, we propose and compare different ways of combining the list of actions from several base models.
To ensure that the resulting parse is well-formed, we design the methods so that the output parse always matches one of the parses produced by ensembling parsers.

{\bf Majority Vote:} We simply select the list which has been predicted by the majority of the individual parsers as the output of the ensemble.  

{\bf Greedy Action:}  
Starting from left (the top-level intent) to right, we pick the action that is predicted by the majority of the parsers, and then discard parsers that do not agree with the majority.
Since the action lists can have different lengths, we pad the shorter lists with dummy actions which are eventually removed. 

{\bf Parser Switch:} We implement an approach similar to the one proposed in~\cite{Henderson99Ensemble}. 
From each parse $i$, we first obtain the \emph{constituent set} $S_i$, where each constituent is a triple (label, start index, end index) extracted from a sub-tree node. For example, the constituent corresponding to the top intent of the parse in Figure~\ref{fig:examples} is $(\texttt{IN:GET\_DISTANCE},0,6)$. The score for each parser $i$ is then computed as $\sum_j |S_i \cap S_j|$; i.e., the sum of intersection with the constituent sets from other parses. Finally, we choose the parser with the highest score. In contrast to the previous two methods, parser switch considers the whole tree to find the majority consensus, but it is computationally more expensive as it needs to construct trees from all parsers to obtain the constituent sets.

We also report the oracle ensemble parser which always picks the correct parser whenever possible in order to determine the upper bound for ensembling gains. For the rest of the paper, we use seven parsers for the ensemble model. The results in Table~\ref{tab:ensemble_results} show that the greedy action and majority vote strategies are almost equally good and better than the parser switch.
On the other hand, all the methods fall short of the oracle parser which suggests that there might still be room for improving the ensemble parser.
The effectiveness of the simple majority vote confirms the observation in~\cite{Surdeanu2010ensemble},
which showed that simple voting performs well for ensembling dependency parsers.

\begin{table}[t]
\begin{minipage}[t]{2.5in}
  \caption{Gains from ensembling.}
  \label{tab:ensemble_results}
  \centering
  \begin{tabular}{lr}
    \toprule
    \cmidrule(r){1-2}
    Ensembling Strategy     & Accuracy      \\
    \midrule
    Base  &  80.86     \\
    Oracle &  90.81 \\
    \midrule
    Majority Vote & 83.76     \\
    Greedy Action    & {\bf 83.84}       \\
    Parser Switch     & 82.82         \\
    \bottomrule
  \end{tabular}
\end{minipage}
    \hfill
\begin{minipage}[t]{2.5in}
  \caption{Gains from ELMo embeddings.}
  \label{tab:elmo_results}
  \centering
  \begin{tabular}{lr}
    \toprule
    \cmidrule(r){1-2}
    ELMo layer     & Accuracy      \\
    \midrule
    First  &  {\bf83.93}      \\
    Last & 83.08     \\
    Average    & 83.85       \\
    Learned     & 83.76         \\
      oncatenated  & 83.70 	\\
    \bottomrule
  \end{tabular}
  \end{minipage}
  
\end{table}

\subsection{Contextual word embeddings}

Following the substantial gains provided by contextual embeddings for multiple NLP tasks~\cite{Peters:2018}, we replace the regular pre-trained word embeddings with off-the-shelf pre-trained ELMo embeddings. It has 2 bi-LSTM layers with highway connections on top of a character n-gram convolutional layer, which after projection result in three vectors of size 1024 for each word.

The results for different ELMo strategies are shown in Table~\ref{tab:elmo_results}: using the first layer, last layer, average of 3 layers, concatenating the layers, and learned weighted average of the layers. The first and last layer weights result in the biggest and smallest gains, respectively. A justification for the first layer's better performance on our task can be that the parser needs to identify syntactic properties (e.g., opening the non-terminal right after prepositions) to produce correct parses. 

\subsection{Language Model Re-ranking}

The experiment in \cite{gupta2018rnng} showed that the top-5 accuracy of the model (i.e., the fraction of utterances where the correct parse is in any of the output from a beam of 5) is much higher than the top-1 accuracy. We train the base model as described in the previous section, but use beam search of size 5 instead of greedy decoding at inference. This improves the accuracy of the top-1 and top-5 hypotheses as shown in Table~\ref{tab:top_k_accuracy}. This suggests that re-ranking the top hypotheses can be an effective way of increasing the accuracy.  

Similar to \cite{Collins:2005:rerank}, we propose to score the parses for re-ranking with a language model over the sequential serialization of the parses.
Our hypothesis is that the generative nature of language model scoring could mitigate some of errors arising from the greedy decoding in RNNG.
We serialize the parse tree as follows: the opening bracket and the non-terminal are considered as one token, and the closing bracket is mapped to the corresponding non-terminal and considered a different token.
For example, \texttt{[IN:GET\_EVENT [SL:CAT\_EVENT} \emph{Concerts} \texttt{]} \emph{by} \texttt{[SL:NAME\_EVENT} \emph{Kendrick Lamar} \texttt{] ]} would become "\texttt{O\_IN\_GET\_EVENT O\_SL\_CATEGORY\_EVENT} \emph{Concerts} \texttt{C\_SL\_CATEGORY\_EVENT} \emph{by} \texttt{O\_SL\_NAME\_EVENT} \emph{Kendrick Lamar} \texttt{C\_SL\_NAME\_EVENT C\_IN\_GET\_EVENT}" (\texttt{O} stands for open and \texttt{C} stands for close).

We serialize the gold trees in the training data and train a neural language model on them. We explore two approaches to re-ranking using the trained language model:
\begin{itemize}
    \item Use the LM score directly to rank the parses. We found that re-ranking only the top two hypotheses gives the best results.
    \item Train a SVM-based ranker~\cite{Herbrich:99} that takes the beam score and LM score as features. 
\end{itemize}

\begin{table}
  \caption{Gains from LM-based re-ranking with beam size 5.}
  \label{tab:top_k_accuracy}
  \centering
  \begin{tabular}{lr}
    \toprule
    \cmidrule(r){1-2}
    Method     & Accuracy      \\
    \midrule
    Base (Greedy) &  80.86      \\
    Oracle top-2 beam     & 89.99        \\
    Oracle top-5 beam     & 93.20         \\
    \midrule
    Top-1 beam     & 81.21     \\
    Naive LM ranker  & 82.80 	\\
    SVM ranker       & {\bf 84.26} \\
    \bottomrule
  \end{tabular}
\end{table}

At the end, the top decoded beams are scored and re-ranked based on one of the two strategies described above. The last two rows in Table~\ref{tab:top_k_accuracy} show that the LM re-ranking using the SVM ranker results in a substantial gain over the top-1 beam but there is still a huge difference compared to the oracle top-5 beam decoding. This suggests that more effective re-ranking approaches, such as including more features in the ranker, could be worth trying.

\section {Error Analysis}

In this section, we categorize the types of errors the parsing model makes in our task and analyze which ones can be mitigated via the methods described in the previous section. We classify the errors into  seven major groups. Note that a single query can exhibit multiple types of errors. (In the list below, -- and + denote the expected and predicted frames below, respectively.)

\begin{enumerate} 
\item { \bf Wrong Top intent (WT):} The first action taken is wrong.

-- \texttt{[IN:GET\_INFO\_TRAFFIC} \emph{ how many miles is } \texttt{[SL:LOCATION} \texttt{[IN:GET\_LOCATION} \texttt{[SL:CATEGORY\_LOCATION} \emph{ the interstate } \texttt{]} \texttt{]} \texttt{]} \emph{ backed up } \texttt{]}

+ \texttt{[\color{red}{IN:GET\_DISTANCE}} \emph{ how many } \texttt{[SL:UNIT\_DISTANCE} \emph{ miles } \texttt{]} \emph{ is } \texttt{[SL:DESTINATION} \texttt{[IN:GET\_LOCATION} \texttt{[SL:CATEGORY\_LOCATION} \emph{ the interstate } \texttt{]} \texttt{]} \texttt{]} \emph{ backed up } \texttt{]}

\item {\bf Wrong Label (WL):} A predicted label for the sub-intents or slots is wrong.

-- \texttt{[IN:GET\_INFO\_TRAFFIC} \emph{ Should I avoid } \texttt{[SL:PATH\_AVOID} \emph{ I - 26 } \texttt{]} \texttt{[SL:DATE\_TIME} \emph{ today } \texttt{]} \texttt{]}

+ \texttt{[IN:GET\_INFO\_TRAFFIC} \emph{ Should I avoid } \texttt{[\color{red}{SL:LOCATION}} \emph{ I - 26 } \texttt{]} \texttt{[SL:DATE\_TIME} \emph{ today } \texttt{]} \texttt{]}

\item {\bf Spurious span (SS):} A Constituent is wrongly added.

-- \texttt{[IN:GET\_ESTIMATED\_DEPARTURE} \emph{Do I need to leave earlier} \texttt{[SL:DATE\_TIME\_DEPARTURE} \emph{ today } \texttt{]} \emph{ due to traffic } \texttt{]}

+ \texttt{[IN:GET\_ESTIMATED\_DEPARTURE} \emph{Do I need to leave} \texttt{\color{red}{[SL:SOURCE} \emph{ earlier } \texttt{]}} \texttt{[SL:DATE\_TIME\_DEPARTURE} \emph{ today } \texttt{]} \emph{ due to traffic } \texttt{]}

\item {\bf Missing Spans (MS): } A Constituent is missing in the prediction.

-- \texttt{[IN:GET\_INFO\_TRAFFIC} \emph{ What is traffic like in } \texttt{[SL:LOCATION} \emph{ still water } \texttt{]} \texttt{[SL:DATE\_TIME} \emph{ at the moment } \texttt{]} \texttt{]}

+ \texttt{[IN:GET\_INFO\_TRAFFIC} \emph{ What is traffic like in \color{red}{still water} } \texttt{[SL:DATE\_TIME} \emph{ at the moment } \texttt{]} \texttt{]}

\item {\bf Wrong Split (WS):} Constituents are wrongly split.

-- \texttt{[IN:GET\_EVENT} \emph{ When is } \texttt{[SL:CATEGORY\_EVENT} \emph{ Christmas in the Park } \texttt{]} \texttt{[SL:DATE\_TIME} \emph{ this year } \texttt{]} \emph{ in } \texttt{[SL:LOCATION} \emph{ San Antonio TX } \texttt{]} \texttt{]}

+ \texttt{[IN:GET\_EVENT} \emph{ When is } \texttt{\color{red}{[SL:DATE\_TIME} \emph{ Christmas } \texttt{]} \emph{ in  \texttt{[SL:LOCATION}} \texttt{[IN:GET\_LOCATION} \texttt{[SL:CATEGORY\_LOCATION} \emph{ the Park } \texttt{]} \texttt{]} \texttt{]}} \texttt{[SL:DATE\_TIME} \emph{ this year } \texttt{]} \emph{ in } \texttt{[SL:LOCATION} \emph{ San Antonio TX } \texttt{]} \texttt{]}

\item {\bf Wrong join (WJ):} Constituents are wrongly joined.

-- \texttt{[IN:GET\_EVENT} \texttt{[SL:LOCATION} \emph{ Dayton } \texttt{]} \texttt{[SL:CATEGORY\_EVENT} \emph{ parties } \texttt{]} \texttt{[SL:DATE\_TIME} \emph{ for NYE } \texttt{]} \texttt{]}

+ \texttt{[IN:GET\_EVENT} \texttt{[SL:CATEGORY\_EVENT} \emph{ \color{red}{Dayton parties} } \texttt{]} \texttt{[SL:DATE\_TIME} \emph{ for NYE } \texttt{]} \texttt{]}

\item{\bf Bad Boundary (BB):} A constituent has wrong boundaries.

-- \texttt{[IN:GET\_EVENT} \emph{ Whats } \texttt{[SL:DATE\_TIME} \emph{ tomorrows } \texttt{]} \emph{ events for } \texttt{[SL:LOCATION} \emph{ Houston } \texttt{]} \texttt{]}

+ \texttt{[IN:GET\_EVENT} \emph{ Whats } \texttt{[SL:CATEGORY\_EVENT} \emph{ \color{red}{tomorrows events} } \texttt{]} \emph{ for } \texttt{[SL:LOCATION} \emph{ Houston } \texttt{]} \texttt{]}
\end{enumerate}

The absolute number of errors for the base model and the relative change for each of the methods in the previous section are summarized in Table~\ref{tab:error_analysis}. We use the majority vote ensemble and the average ELMo setting for the experiments. We can see that ELMo is effective across all types of errors, while ensemble and LM ranking mitigate most types of errors. Ensembling is much more effective at top intent classification (a semantic task) whereas ELMo is the only method effective against missing span and wrong join (mostly syntactic tasks). This confirms the hypothesis that the RNNG model uses the syntactic information in the first layer of ELMo more than the other layers. The above analysis suggests that combining the aforementioned methods can be effective in mitigating different types of errors together, which we explore in the next section.

\begin{table}
  \caption{Effectiveness of methods on different types of errors (as percentage of error reduction).}
  \label{tab:error_analysis}
  \centering
  \begin{tabular}{lrrrrrrr}
    \toprule
    \cmidrule(r){1-2}
    Method     & WT  & WL & MS  & SS &  WS   & WJ & BB \\ %
    \midrule
    Base  		&  222  & 850  & 246  &  333 & 110  & 88  & 330  \\ %
    \midrule
    ELMo     	& -9\% 	&  -20\%  & {\bf-11\%}  &  -25\%	 & -22\% 		& {\bf-28\%} & -21\%  \\   %
    Ensemble     & {\bf-25\%}   &-21\%  & +2\% 	& -32\% 		& -29\%  		& +3\% 	& -18\%  \\ %
     LM re-rank     &  -20\% 	&-20\%    	& +3\% 	& {\bf -36}\% 		& {\bf-47}\% 		& -8\%  	& -15\% \\ %
    \bottomrule
  \end{tabular}
\end{table}

\section {Combining the Methods}
Based on the analysis in the previous section, we combine the methods to seek further gains. We first try the ELMo + Ensemble combination. We use the average ELMo strategy and experiment with the three ensemble strategies. The results are shown in Table~\ref{tab:elmo_ensemble_results}. The combined gain confirms that the ensemble and ELMo work almost orthogonally and the combination results in about $28\%$ error reduction compared to the base parser.

\begin{table}
\begin{minipage}[b]{2.5in}
  \caption{Combining ensembling and ELMo.}
  \label{tab:elmo_ensemble_results}
  \centering
  \begin{tabular}{lr}
    \toprule
    \cmidrule(r){1-2}
    Ensemble Strategy   & Accuracy      \\
    \midrule
    Oracle &  92.43 \\
    \midrule
    Majority Vote & {\bf 86.19}    \\
    Greedy Action    & 85.95    \\
    Parser Switch     & 84.62        \\
    \bottomrule
  \end{tabular}
\end{minipage}
\hfill
\begin{minipage}[b]{2.5in}
\caption{LM re-ranking with ELMo.}
  \label{tab:elmo_lm_results}
  \centering
  \begin{tabular}{lr}
    \toprule
    \cmidrule(r){1-2}
    Method     & Accuracy      \\
    \midrule
    Oracle top-2 beam     & 91.53         \\
    Oracle top-5 beam     & 95.04        \\
    \midrule
    Base ELMo (Greedy) &  83.85      \\
    Top-1 beam     &   84.05 \\
    LM re-ranked  & {\bf 86.30}	\\
    \bottomrule
  \end{tabular}
  \end{minipage}
\end{table}

We also experiment with adding the LM SVM re-ranking to the ELMo-enabled model. The results are shown in Table~\ref{tab:elmo_lm_results}. We observe that combined with ELMo, the LM re-ranking of the top beams still keeps most of its gains compared to the case without ELMo.

 Finally, we report the results of combining LM re-ranking and ensemble (with and without using ELMo). Here, we employ two strategies. In the first strategy, using the ranking SVM strategy as before, we re-rank the top five hypotheses for each parser inside the ensemble and then apply the ensemble strategy. In the second strategy, we use the number of times each hypothesis appears in the top-5 beam of the parsers as an additional feature to the ranking SVM.   We use the average ELMo strategy alongside with the majority vote ensemble for these experiments. 
 
 Table~\ref{tab:lm_ensemble_results} compares the results to the baseline using the top beam. We can see that LM re-ranking can be effective on top of ensemble and ELMo, and the biggest gains are achieved by using the voting decision inside the ranking SVM.
  
 \begin{table}
  \caption{The results from combining all techniques.}
  \label{tab:lm_ensemble_results}
  \centering
  \begin{tabular}{lr}
    \toprule
    \cmidrule(r){1-2}
    Method     & Accuracy      \\
    \midrule
    Base Ensemble w/o ELMo  & 83.83 \\
    LM re-ranked Ensemble w/o ELMo     & 84.86        \\
    Extended SVM Ranking w/o ELMo & 85.22 \\
    Base Ensemble with ELMo  & 86.26 \\
    LM re-ranked Ensemble with ELMo    &  86.67      \\
    Extended SVM Ranking with ELMo & \bf{87.25} \\
    \bottomrule
  \end{tabular}
\end{table}

\section{Related Work}
Our work builds on top of two related but distinct directions of research. At one end, there has been a large literature on language understanding for task oriented dialog, such as the work that tackle the ATIS and DSTC datasets~\cite{mesnil2013,Liu2016AttentionBasedRN}. Most work in this area assumes that the utterance is not compositional. The current state-of-the-art \cite{7953243} frames the problem as one of non-recursive intent and slot tagging, and assumes that the NLU output is passed along to a dialog manager in order to be executed. There has also been work on end-to-end task oriented dialog~\cite{bordes-goal:16}, but there too, the problem is usually framed as one of selecting a single API call and its arguments, as opposed to compositional API calls.

At the other end of the spectrum, in the traditional semantic parsing literature, the problem is framed as predicting compositional semantic representations.
These are mainly geared towards question-answering rather than task completion, and are usually directly executed against a knowledge base.
Some of the standard datasets in this area include GeoQuery \cite{zelle1996learning} and WebQuestions \cite{berant2013semantic}.

Within both of these areas, neural approaches have supplanted previous feature-engineering based approaches in recent years \cite{HakkaniTur2016,IyerKCKZ17}.
In the context of tree-structured semantic parsing, some other interesting approaches include Seq2Tree~\cite{DBLP:journals/corr/DongL16} which modifies the standard Seq2Seq decoder to better output trees; SCANNER~\cite{DBLP:journals/corr/abs-1711-05066,DBLP:journals/corr/0001RSL17} which extends the RNNG formulation specifically for semantic parsing such that the output is no longer coupled with the input; and TRANX \cite{tranx} and Abstract Syntax Network \cite{thatHearthstonePaper} which generate code along a programming language schema. For graph-structured semantic parsing  \cite{banarescu2013amr,he2017deep}, SLING~\cite{DBLP:journals/corr/abs-1710-07032} produces graph-structured parses by modeling semantic parsing as a neural transition parsing problem with a more expressive transition tag set. While graph structures can provide more detailed semantics, they are more difficult to parse and can be an overkill for understanding task oriented utterances. 

\section{Conclusion}
In this paper, we propose three different techniques to improve the hierarchical representation based semantic parsing models using ensembling, contextualized word embeddings, and language model based re-ranking. We propose a categorization of errors for the TOP representation. Our results show that the three approaches improve the model on different types of errors and the best model uses a combination of them. Our best model reduces the error rate by 33\% on the TOP dataset.

\bibliographystyle{abbrvnat}
\bibliography{refs}

\begin{thebibliography}{25}
\providecommand{\natexlab}[1]{#1}
\providecommand{\url}[1]{\texttt{#1}}
\expandafter\ifx\csname urlstyle\endcsname\relax
  \providecommand{\doi}[1]{doi: #1}\else
  \providecommand{\doi}{doi: \begingroup \urlstyle{rm}\Url}\fi

\bibitem[Banarescu et~al.(2013)Banarescu, Bonial, Cai, Georgescu, Griffitt,
  Hermjakob, Knight, Koehn, Palmer, and Schneider]{banarescu2013amr}
L.~Banarescu, C.~Bonial, S.~Cai, M.~Georgescu, K.~Griffitt, U.~Hermjakob,
  K.~Knight, P.~Koehn, M.~Palmer, and N.~Schneider.
\newblock Abstract meaning representation for sembanking.
\newblock In \emph{7th Linguistic Annotation Workshop and Interoperability with
  Discourse}, 2013.

\bibitem[Berant et~al.(2013)Berant, Chou, Frostig, and
  Liang]{berant2013semantic}
J.~Berant, A.~Chou, R.~Frostig, and P.~Liang.
\newblock Semantic parsing on freebase from question-answer pairs.
\newblock In \emph{Proceedings of the 2013 Conference on Empirical Methods in
  Natural Language Processing}, pages 1533--1544, 2013.

\bibitem[Bordes et~al.(2016)Bordes, Boureau, and Weston]{bordes-goal:16}
A.~Bordes, Y.-L. Boureau, and J.~Weston.
\newblock Learning end-to-end goal-oriented dialog.
\newblock In \emph{Proc. of ICLR}, 2016.

\bibitem[Cheng et~al.(2017{\natexlab{a}})Cheng, Reddy, Saraswat, and
  Lapata]{DBLP:journals/corr/0001RSL17}
J.~Cheng, S.~Reddy, V.~Saraswat, and M.~Lapata.
\newblock Learning structured natural language representations for semantic
  parsing.
\newblock In \emph{Proceedings of the 55th Annual Meeting of the Association
  for Computational Linguistics (Volume 1: Long Papers)}, pages 44--55.
  Association for Computational Linguistics, 2017{\natexlab{a}}.
\newblock \doi{10.18653/v1/P17-1005}.
\newblock URL \url{http://www.aclweb.org/anthology/P17-1005}.

\bibitem[Cheng et~al.(2017{\natexlab{b}})Cheng, Reddy, Saraswat, and
  Lapata]{DBLP:journals/corr/abs-1711-05066}
J.~Cheng, S.~Reddy, V.~Saraswat, and M.~Lapata.
\newblock Learning an executable neural semantic parser.
\newblock \emph{CoRR}, abs/1711.05066, 2017{\natexlab{b}}.
\newblock URL \url{http://arxiv.org/abs/1711.05066}.

\bibitem[Collins and Koo(2005)]{Collins:2005:rerank}
M.~Collins and T.~Koo.
\newblock Discriminative reranking for natural language parsing.
\newblock \emph{Comput. Linguist.}, 31\penalty0 (1):\penalty0 25--70, Mar.
  2005.
\newblock ISSN 0891-2017.
\newblock \doi{10.1162/0891201053630273}.
\newblock URL \url{http://dx.doi.org/10.1162/0891201053630273}.

\bibitem[Dong and Lapata(2016)]{DBLP:journals/corr/DongL16}
L.~Dong and M.~Lapata.
\newblock Language to logical form with neural attention.
\newblock In \emph{Proceedings of the 54th Annual Meeting of the Association
  for Computational Linguistics (Volume 1: Long Papers)}, pages 33--43, Berlin,
  Germany, August 2016. Association for Computational Linguistics.
\newblock URL \url{http://www.aclweb.org/anthology/P16-1004}.

\bibitem[Dyer et~al.(2016)Dyer, Kuncoro, Ballesteros, and Smith]{dyer-rnng:16}
C.~Dyer, A.~Kuncoro, M.~Ballesteros, and N.~A. Smith.
\newblock Recurrent neural network grammars.
\newblock In \emph{Proc. of NAACL}, 2016.

\bibitem[Gupta et~al.(2018)Gupta, Shah, Mohit, Kumar, and Lewis]{gupta2018rnng}
S.~Gupta, R.~Shah, M.~Mohit, A.~Kumar, and M.~Lewis.
\newblock Semantic parsing for task oriented dialog using hierarchical
  representations.
\newblock In \emph{Proceedings of the 2018 Conference on Empirical Methods in
  Natural Language Processing (EMNLP)}, 2018.

\bibitem[Hakkani-Tur et~al.(2016)Hakkani-Tur, Tur, Celikyilmaz, Chen, Gao,
  Deng, and Wang]{HakkaniTur2016}
D.~Hakkani-Tur, G.~Tur, A.~Celikyilmaz, Y.-N. Chen, J.~Gao, L.~Deng, and Y.-Y.
  Wang.
\newblock Multi-domain joint semantic frame parsing using bi-directional
  {RNN-LSTM}.
\newblock In \emph{Interspeech 2016}, pages 715--719, 2016.
\newblock \doi{10.21437/Interspeech.2016-402}.
\newblock URL \url{http://dx.doi.org/10.21437/Interspeech.2016-402}.

\bibitem[He et~al.(2017)He, Lee, Lewis, and Zettlemoyer]{he2017deep}
L.~He, K.~Lee, M.~Lewis, and L.~Zettlemoyer.
\newblock Deep semantic role labeling: What works and what’s next.
\newblock In \emph{Proceedings of the Annual Meeting of the Association for
  Computational Linguistics}, 2017.

\bibitem[Henderson and Brill(2000)]{Henderson99Ensemble}
J.~C. Henderson and E.~Brill.
\newblock Exploiting diversity in natural language processing: Combining
  parsers.
\newblock \emph{CoRR}, cs.CL/0006003, 2000.
\newblock URL \url{http://arxiv.org/abs/cs.CL/0006003}.

\bibitem[Herbrich et~al.(1999)Herbrich, Graepel, and Obermayer]{Herbrich:99}
R.~Herbrich, T.~Graepel, and K.~Obermayer.
\newblock Support vector learning for ordinal regression.
\newblock In \emph{1999 Ninth International Conference on Artificial Neural
  Networks ICANN 99. (Conf. Publ. No. 470)}, volume~1, pages 97--102 vol.1,
  Sept 1999.
\newblock \doi{10.1049/cp:19991091}.

\bibitem[Iyer et~al.(2017)Iyer, Konstas, Cheung, Krishnamurthy, and
  Zettlemoyer]{IyerKCKZ17}
S.~Iyer, I.~Konstas, A.~Cheung, J.~Krishnamurthy, and L.~Zettlemoyer.
\newblock Learning a neural semantic parser from user feedback.
\newblock In \emph{Proceedings of the 55th Annual Meeting of the Association
  for Computational Linguistics, {ACL} 2017, Vancouver, Canada, July 30 -
  August 4, Volume 1: Long Papers}, pages 963--973, 2017.
\newblock \doi{10.18653/v1/P17-1089}.
\newblock URL \url{https://doi.org/10.18653/v1/P17-1089}.

\bibitem[Kingma and Ba(2015)]{kingma2015adam}
D.~P. Kingma and J.~Ba.
\newblock Adam: A method for stochastic optimization.
\newblock In \emph{International Conference for Learning Representations},
  2015.

\bibitem[Liu and Lane(2016)]{Liu2016AttentionBasedRN}
B.~Liu and I.~Lane.
\newblock Attention-based recurrent neural network models for joint intent
  detection and slot filling.
\newblock In \emph{INTERSPEECH}, 2016.

\bibitem[Mesnil et~al.(2013)Mesnil, He, Deng, and Bengio]{mesnil2013}
G.~Mesnil, X.~He, L.~Deng, and Y.~Bengio.
\newblock Investigation of recurrent-neural-network architectures and learning
  methods for spoken language understanding.
\newblock In \emph{Interspeech}, 2013.

\bibitem[Peters et~al.(2018)Peters, Neumann, Iyyer, Gardner, Clark, Lee, and
  Zettlemoyer]{Peters:2018}
M.~E. Peters, M.~Neumann, M.~Iyyer, M.~Gardner, C.~Clark, K.~Lee, and
  L.~Zettlemoyer.
\newblock Deep contextualized word representations.
\newblock In \emph{Proc. of NAACL}, 2018.

\bibitem[Rabinovich et~al.(2017)Rabinovich, Stern, and
  Klein]{thatHearthstonePaper}
M.~Rabinovich, M.~Stern, and D.~Klein.
\newblock Abstract syntax networks for code generation and semantic parsing.
\newblock In \emph{Annual Meeting of the Association for Computational
  Linguistics (ACL)}, 2017.

\bibitem[Recht et~al.(2011)Recht, Re, Wright, and Niu]{recht2011hogwild}
B.~Recht, C.~Re, S.~Wright, and F.~Niu.
\newblock Hogwild: A lock-free approach to parallelizing stochastic gradient
  descent.
\newblock In \emph{Advances in Neural Information Processing Systems (NIPS)},
  2011.

\bibitem[Ringgaard et~al.(2017)Ringgaard, Gupta, and
  Pereira]{DBLP:journals/corr/abs-1710-07032}
M.~Ringgaard, R.~Gupta, and F.~C.~N. Pereira.
\newblock {SLING:} {A} framework for frame semantic parsing.
\newblock \emph{CoRR}, abs/1710.07032, 2017.
\newblock URL \url{http://arxiv.org/abs/1710.07032}.

\bibitem[Surdeanu and Manning(2010)]{Surdeanu2010ensemble}
M.~Surdeanu and C.~D. Manning.
\newblock Ensemble models for dependency parsing: Cheap and good?
\newblock In \emph{Human Language Technologies: The 2010 Annual Conference of
  the North American Chapter of the Association for Computational Linguistics},
  HLT '10, pages 649--652, Stroudsburg, PA, USA, 2010. Association for
  Computational Linguistics.

\bibitem[Yin and Neubig(2018)]{tranx}
P.~Yin and G.~Neubig.
\newblock Tranx: A transition-based neural abstract syntax parser for semantic
  parsing and code generation.
\newblock In \emph{Proceedings of the 2018 Conference on Empirical Methods in
  Natural Language Processing: System Demonstrations}, pages 7--12. Association
  for Computational Linguistics, 2018.

\bibitem[Zelle and Mooney(1996)]{zelle1996learning}
J.~M. Zelle and R.~J. Mooney.
\newblock Learning to parse database queries using inductive logic programming.
\newblock In \emph{Proceedings of the Thirteenth National Conference on
  Artificial Intelligence (AAAI)}, 1996.

\bibitem[Zhu and Yu(2017)]{7953243}
S.~Zhu and K.~Yu.
\newblock Encoder-decoder with focus-mechanism for sequence labelling based
  spoken language understanding.
\newblock In \emph{2017 IEEE International Conference on Acoustics, Speech and
  Signal Processing (ICASSP)}, pages 5675--5679, March 2017.
\newblock \doi{10.1109/ICASSP.2017.7953243}.

\end{thebibliography}

\end{document}